\documentclass[letterpaper, 10pt, conference]{template/ieeeconf}    

\IEEEoverridecommandlockouts    

\usepackage{graphicx} 
\usepackage{mathptmx} 
\usepackage{times} 
\usepackage{amsmath} 
\usepackage{amssymb}  
\usepackage{subcaption} 
\usepackage{booktabs} 
\usepackage{multirow} 
\usepackage{hyperref}
\usepackage{siunitx}

\usepackage[dvipsnames]{xcolor}
\usepackage[switch]{lineno}

\graphicspath{ {./figures/} }

\title{\LARGE \bf{Fast Trajectory End-Point Prediction with Event Cameras\protect\\ for Reactive Robot Control}}

\author{{Marco Monforte, Luna Gava, Massimiliano Iacono, Arren Glover, and Chiara Bartolozzi} 
\thanks{{All the authors are with the Event-Driven Perception for Robotics Research Line, Istituto Italiano di Tecnologia, Italy. \{\tt\small marco.monforte, luna.gava, massimiliano.iacono, arren.glover, chiara.bartolozzi\}@iit.it}}}

\begin{document}

\maketitle


\begin{abstract}
Prediction skills can be crucial for the success of tasks where robots have limited time to act or joints actuation power. In such a scenario, a vision system with a fixed, possibly too low, sampling rate could lead to the loss of informative points, slowing down prediction convergence and reducing the accuracy. In this paper, we propose to exploit the low latency, motion-driven sampling, and data compression properties of event cameras to overcome these issues. 
As a use-case, we use a Panda robotic arm to intercept a ball bouncing on a table. To predict the interception point, we adopt a Stateful LSTM network, a specific LSTM variant without fixed input length, which perfectly suits the event-driven paradigm and the problem at hand, where the length of the trajectory is not defined. We train the network in simulation to speed up the dataset acquisition and then fine-tune the models on real trajectories. Experimental results demonstrate how using a dense spatial sampling (i.e. event cameras) significantly increases the number of intercepted trajectories as compared to a fixed temporal sampling (i.e. frame-based cameras).
\end{abstract}



\section{Introduction}
\label{sec:introduction}


Human interaction with the environment strongly benefits from predictive capabilities~\cite{Bar2009, Roach2011, Schacter2008}. 
Similarly, endowing robots with the ability to estimate future outcomes of movements of agents and objects in the environment can improve their performance, in terms of robustness, reliability, and precision. 
When the task has highly varying dynamics or short temporal duration, for instance, the robot has a narrow time window to act, and aspects like the operating rate and actuation limitations become crucial for success. 
A fast prediction could allow to plan actions in advance, minimizing accelerations and compensating for internal delays, or to wait and gather as much information as possible before moving at full speed, according to the robot possibilities, opening new scenarios for robots that do not mount high-quality, high-precision, costly components. 

Event cameras transmit information in response to the motion of edges in their field of view, corresponding to an adaptive sampling tied to the stimulus dynamics, rather than on a fixed independent clock. The information transmitted is inherently sparse and asynchronous, allowing for energy saving, avoiding data redundancy, and yielding sub-millisecond latency~\cite{Posch2011}. This new type of information encoding makes event cameras more suitable than frame-based systems for real-world, high update rate applications, like tracking and prediction~\cite{Gallego2022}. 

\begin{figure}
    \centering
    \begin{subfigure}{.95\linewidth}
        \includegraphics[width=\textwidth]{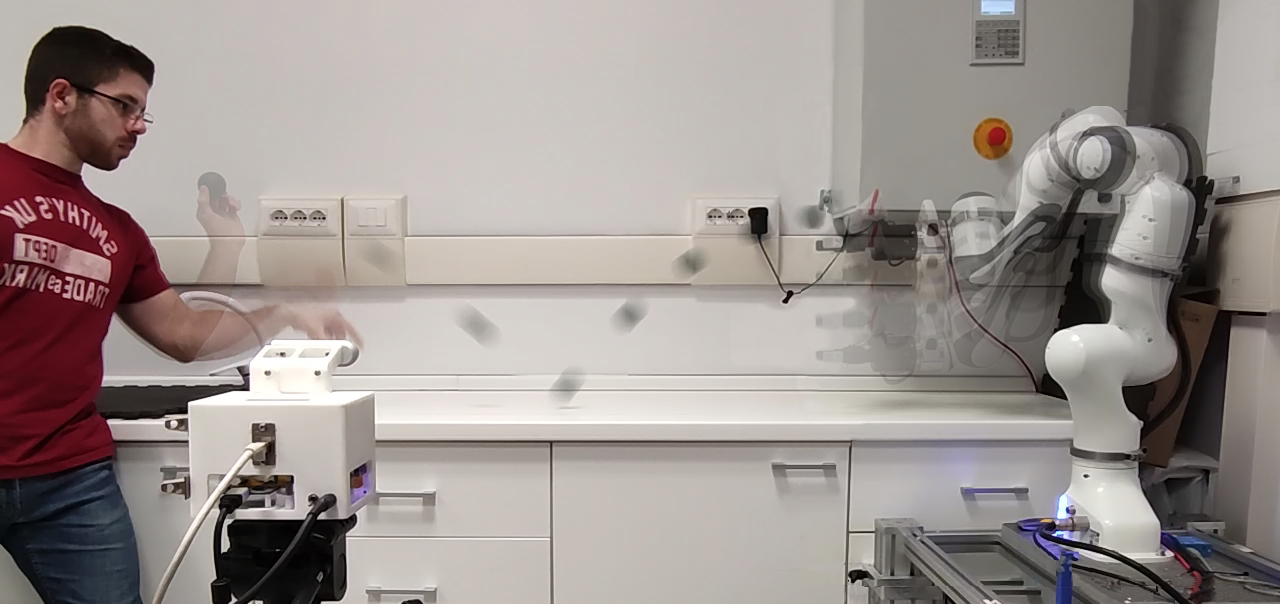}
        \caption{}
        \label{subfig:setup}
    \end{subfigure}
    \begin{subfigure}{.95\linewidth}
        \includegraphics[width=\textwidth]{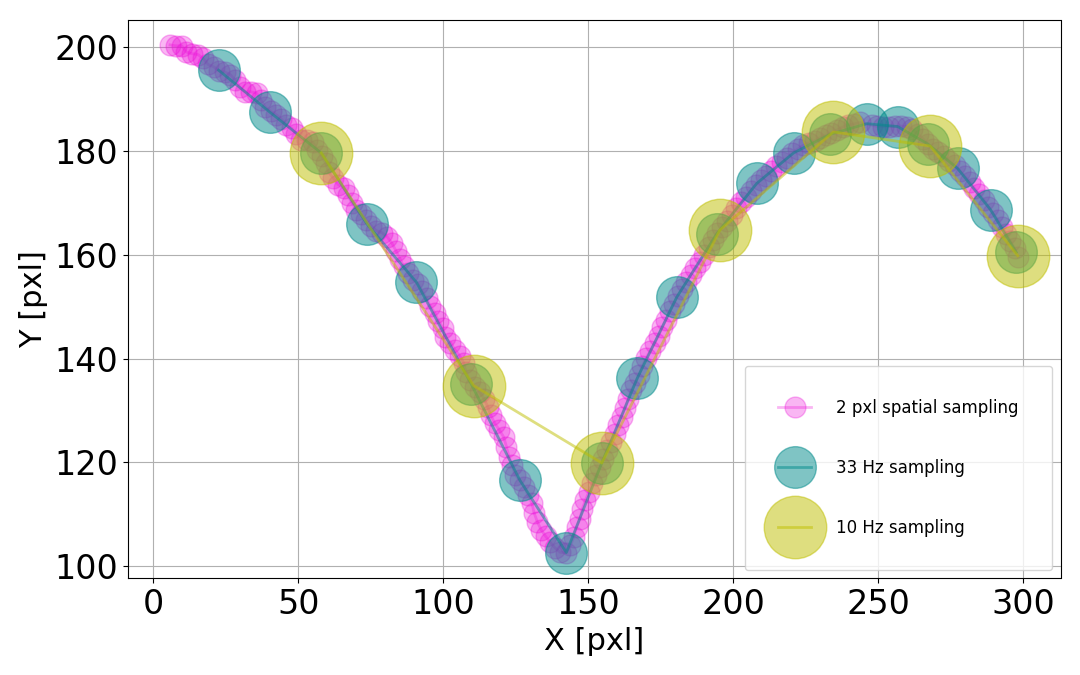}
        \caption{}
        \label{subfig:sampling_rates}
    \end{subfigure}
  \caption{Robotic arm intercepting a bouncing ball. (a) Setup. (B) Validation: We compare three different sampling strategies - asynchronous spatial event-driven (the target position is sampled every time it moves of 2 pixels) and synchronous temporal 33 Hz and 10 Hz.}
  \label{fig:setup_and_rates}
\end{figure}

Prediction tasks have been addressed with a wide variety of approaches in the mainstream frame-based robotics literature. 
The catching of a ball in flight with a robotic arm has been tackled using a stereo method to track its 3D position and an Extended Kalman Filter (EKF) for predicting future states through numerical integration for a finite horizon~\cite{Sato2020, Dong2020}. A combination of Support Vector Regression and EKF tracks markers on an object and predicts up to \SI{1}{s} ahead at \SI{200}{Hz}~\cite{Kim2014}, enabling grasping of objects in flight. A drawback of EKF methods, however, is that all the predicted steps from the current instant must be recomputed every time a new measure arrives.
In place of EKF or other model-based methods, data-driven methods like neural networks have been used in~\cite{Zhao2019a, Asenov2020}. Zhao et al.~\cite{Zhao2019a} address the human-robot handover task, highlighting the importance of a well-timed robot movement for a person to perceive it as human-like. A Long Short-Term Memory (LSTM)~\cite{Hochreiter1997a} network predicts what the robot's next joints' configuration should be in response to the human motion. Results show the system successfully adapts the time scale of the motion to the giver's one.
In~\cite{Chung2015} a Variational Recurrent Neural Network (VRNN) combined with a dynamic model of the task is capable of predicting the trajectory end-point of a ball bouncing on a table in the 2D frontal plane of a robot, which has to intercept it on the vertical axis at the end of its visual space~\cite{Asenov2020}.
The whole system is trained end-to-end in simulation and deployed in the real world. The fast convergence of the online estimated parameters allows probabilistic predictions even in case of blind spots along the trajectory. A single experiment lasts for \SI{2}-\SI{3}{s}, with the system running at \SI{20}{Hz} and refining the estimate frame after frame. 
All these works make use of powerful, external, and energy consuming devices - like GPUs in~\cite{Zhao2019a}, the VICON motion capture system in~\cite{Dong2020}, or the IDP Express RF2000F system and the ad-hoc finger cameras in~\cite{Sato2020}. This is not always possible, though. Problems arise with autonomous systems having limited battery, or in outdoor tasks where external cameras cannot be placed.

The role of perception on agents actions' timing highlighted how the choice of the right sensing device becomes crucial for the success of a task~\cite{Falanga2019}. 
Event cameras can leverage on very high temporal resolution (for fast moving targets), very high dynamic range ($\sim$\SI{140}{dB})~\cite{Posch2011}, and yet with low energy requirements. 
He et al.~\cite{He2021} propose a low latency, high precision pipeline for dynamic object avoidance with a quadrotor by fusing IMU data, depth, and events. The object's 3D trajectory is predicted by combining events and depth information to obtain the object's location in the depth image, and then estimated with a second-order polynomial. After an offline evaluation of the accuracy, online experiments show the quadrotor avoiding the ball under different scenarios. 
In~\cite{monforte2019} an event camera with an Encoder-Decoder~\cite{Sutskever2014} LSTM network predicts the giver's motion during a handover task from the iCub robot perspective. The pipeline predicts both spatial and temporal future points, allowing the robot to know in advance where to move, compensating for internal delays in the perception-action loop. Running as fast as \SI{250}{Hz}, the predictor immediately perceives changes in the action and adapts its horizon to the giver's motion. 
In~\cite{Huang2022} the authors propose a grasping framework for eye-in-hand robotic manipulators endowed with event cameras, using both model-based and model-free multi-object grasping in clutter. While the former ensures higher precision, the latter is more general and applicable to real-world scenarios. 
Wang et. al~\cite{Wang2022} implemented a system for catching balls thrown by a tennis ball launcher with a 1 degree-of-freedom linear actuator. The proposed Binary Event History Image (BEHI) accumulates information from events in images that are fed to an impact prediction pipeline that controls the actuator motion timing. The actuator catches balls up to a top speed of 13m/s with a success rate of 80\%.

In a similar task, we investigate the advantages of event-driven, asynchronous sampling with respect to a standard frame-based sampling, for predicting a bouncing ball's spatio-temporal location~\cite{Asenov2020}, and we investigate two different possible control strategies a robotic arm might adopt to catch it.
The system predicts the spatial and temporal position of the last visible instance of the ball, hence the last $(x,y)$ spatial coordinates and timestamp, to be intercepted by the robot - Fig.~\ref{subfig:setup}, in the shortest possible time. 
To isolate the specific role of the event cameras and fast prediction without other confounding factors, we make the following assumptions: (i) the ball is the only object moving in the scene and the camera is not attached to the robot; (ii) the trajectory of the ball is approximately planar and perpendicular to the camera plane (i.e. the system performance does not depend on the stereo estimation); and (iii) the robot has a single open-loop attempt to hit the ball (i.e. we only rely on the most up to date prediction at the time the robot starts moving).  We compare the performance of the system using event cameras with 2 pixels spatial sampling, or fixed temporal sampling at 33 and 10 Hertz (as shown in Fig.~\ref{subfig:sampling_rates}), or frame-based cameras at 30 and 60 Hz.
Differently from the model-based approaches typically used in the case of parabolic trajectories, where the physical model of the object trajectory is learnt~\cite{Sato2020, Dong2020, Kim2014} and its parameters are estimated~\cite{Asenov2020}, we use a model-free \textit{Stateful} LSTM neural network to learn from a dataset of trajectories their ending points. The advantage of this architecture is that it does not have a fixed, a priori-defined input length, but instead keeps memory of past information indefinitely and updates its prediction using the latest input. This suits the event-driven paradigm, for which the number of data points cannot be defined before the action unfolds. This ``continuous memory'' also allows feeding as input only the latest point to update the prediction at each time step, avoiding redundant computation in the network due to possible buffers~\cite{monforte2019}. To boost the learning process, and cope with the data-hungry nature of the chosen model-free approach, similarly to~\cite{Asenov2020}, we train on simulated trajectories, parametrized to capture the same statistics of the real ones, before fine-tuning the models in real-world experiments.
Finally, we consider two possible control strategies for the robotic arm: the first is to move as soon as the prediction converges, to minimize the velocity required to perform the movement; the second is to wait as long as possible, accumulating information to reduce the prediction error and move as fast as the robot can. 

With respect to previous work in~\cite{monforte2019, monforte2020}, we:
\begin{itemize}
    \item adopt a Stateful LSTM neural network as the predictor, an architecture dealing with temporal information that has no limits in the input sequence length and does not require any buffer;
    \item make use of simulated data to speed up the dataset acquisition and training process, before refining the obtained models on a real dataset;
    \item validate the approach with real-world experiments with a Panda robotic arm, mapping from pixels to Cartesian coordinates through a visual calibration process;
    \item quantitatively compare the performance of the system using event cameras with asynchronous sampling and frame-like temporal sampling and with frame-based cameras.
\end{itemize}
The use of event cameras enables the tracking of faster targets and improves prediction speed, accuracy and success rate of ball catching.

\section{Methodology}
\label{sec:methodology}


The architecture comprises an ATIS~\cite{Posch2011} event camera placed near the robot, an event-driven tracker~\cite{monforte2020} that collects raw events from the ATIS to localize the target and track its center of mass, a Stateful LSTM network~\cite{Hochreiter1997a} predicting where and when the ball will be out of the camera field of view, updating its estimate every new tracker output received, and a Franka Emika Panda robot manipulator~\cite{Panda}.
The event-driven tracker~\cite{monforte2020} outputs the center of mass position of the target every time this moves. An initial check on the collected events' distribution locates the object and initialise the tracker. Then, a Region Of Interest (ROI) accumulates a user-defined number N of events to be used to continuously update the center of mass of the object.

The tracker runs at $\sim1$ kHz, outputting information only when the center of mass of the target moves, i.e. spatially sampling the trajectory. To reduce the effect of noise on what would be a densely sampled trajectory and lower the overall computational cost, a \textit{spatial sampling} of 2 pixels is performed, i.e. tracker outputs an event only if the new center of mass position is farther than 2 pixels from the previous one. This operation allows to lower the amount of processing required in the following stages of the pipeline, without undermining the resolution of the sampled trajectory nor the accuracy of the future prediction~\cite{monforte2020}. 

Differently from Multi-Layer Perceptrons (MLP) and Convolutional Neural Networks (CNNs), which do not maintain any temporal correlation between consecutive queries, Recurrent Neural Networks (RNNs) inherently do so thanks to recurrent connections, that allow the network computational nodes to re-use information coming from previous queries. 
Long Short-Term Memory networks~\cite{Hochreiter1997a} represent the state-of-the-art in this family of architectures, and are suitable for event-driven frameworks, where the time interval between consecutive updates is variable, and not fixed a priori. On the one hand, this increases the parameter space of the architecture; on the other hand, it explicitly represents the dynamics of the scene, possibly allowing the network to earlier discriminate among the possible solutions. Depending on how many pixels will be triggered along the motion, however, it is not possible to define a narrow range of trajectory lengths. This goes against the standard use of LSTM networks, usually designed to receive a fixed, constant number of input data points. 

We therefore resorted to a \textit{Stateful} LSTM network, a variant of the classical LSTM network that runs indefinitely, without any limit on the number of inputs. The network memory is reset, at run-time, given specific conditions set by the user, in order to analyse a new input sequence. In our application, the memory is reset when no tracker update has been received in the last 2 seconds (i.e. the previous trajectory has ended and the target is not detected).

The Panda robot~\cite{Panda} is a 7-degrees-of-freedom manipulator. Given the actuation limits reported for the robot, intermediate points between the initial and final position are generated with a quintic polynomial, to limit both the initial and final velocity and acceleration. Under such limits, the vertical range assumed for the task is of \SI{60}{cm}, viable in \SI{1.1}{s}. The calibration procedure to map pixel coordinates to the robot Cartesian position has seen the 10 times acquisition of 8 pairs of \textit{(pixel coordinates, robot height)} in this range, before running a quadratic regression to obtain the mapping function. To be considered intercepted, the ball has to be hit by the gripper, which has a height of \SI{2}{cm}.

\subsection{Simulated dataset generation}
\label{subsec:simulated_dataset_generation}
The Unreal Engine environment~\cite{Unreal_Engine} (UE) has been used to generate a total of 5400 synthetic trajectories (5000 for training, 250 for validation, and 150 for testing). Fine-tuning of the UE simulation parameters allows to match and extend the statistics of the simulated data to those of a set of 310 real trajectories 
acquired from the setup shown in Fig.~\ref{subfig:setup}.
Each trajectory comprises a sequence of RGB frames generated by UE at \SI{500}{Hz}, the corresponding event stream generated with ESIM events simulator~\cite{Rebecq18corl}, and the output of the event-driven tracker ($(x,y)$ spatial coordinates and timestamp of the ball's center of mass).

\subsection{Prediction convergence criteria and parameters}
At run-time, there is no ground-truth with which to compare the system estimate and evaluate if the error is small enough to plan and execute the action of the robot. We define, therefore, the parameter $\gamma$ to determine when the prediction converged and the robot can move. This parameter is based on the rate of change of the previous N estimates of the final vertical position of the target $y_F$:
\begin{equation}
\gamma(i) = \frac{1}{N} \sum_{j=i-N+1}^{i} \frac{|\hat{y}_F(j) - \hat{y}_F(j-1)|}{t(j) - t(j-1)}
\label{eq:gamma}
\end{equation}
where $t_i$ is the time at the $i$-ith point, with $i=N,\dots,M$, N depending on the sampling strategy and M being the total number of samples for the current trajectory. 
In such a way, we can define the \textit{convergence instant $t_{conv}$} as the first moment such that $\gamma (i)$ is below a user-defined threshold $\gamma^*$:
\begin{equation}
t_{conv} = t_1 \in T=\{t(i) | \gamma(i) < \gamma^*\}
\label{eq:t_conv}
\end{equation}
with $t_1$ being the first element of the set of time instants T where the condition on $\gamma$ is satisfied.
Moreover, knowing the average robot velocity $v^{robot}$ and its starting position $y^{robot}_{start}$, we can compute the time needed to reach any point in the task space. This value, along with the final time $\hat{t}_F$, is used to compute the last moment the robot can wait before moving $\hat{t}_{dec}$, to reach $\hat{y}_F$:
\begin{equation}
\hat{t}_{dec}(i) = \hat{t}_F(i) - \frac{y_{start}^{robot} - \hat{y}_F(i)}{v^{robot}}
\label{eq:t_dec}
\end{equation}

\begin{figure}[t]
    \centering
    \includegraphics[width=0.49\textwidth]{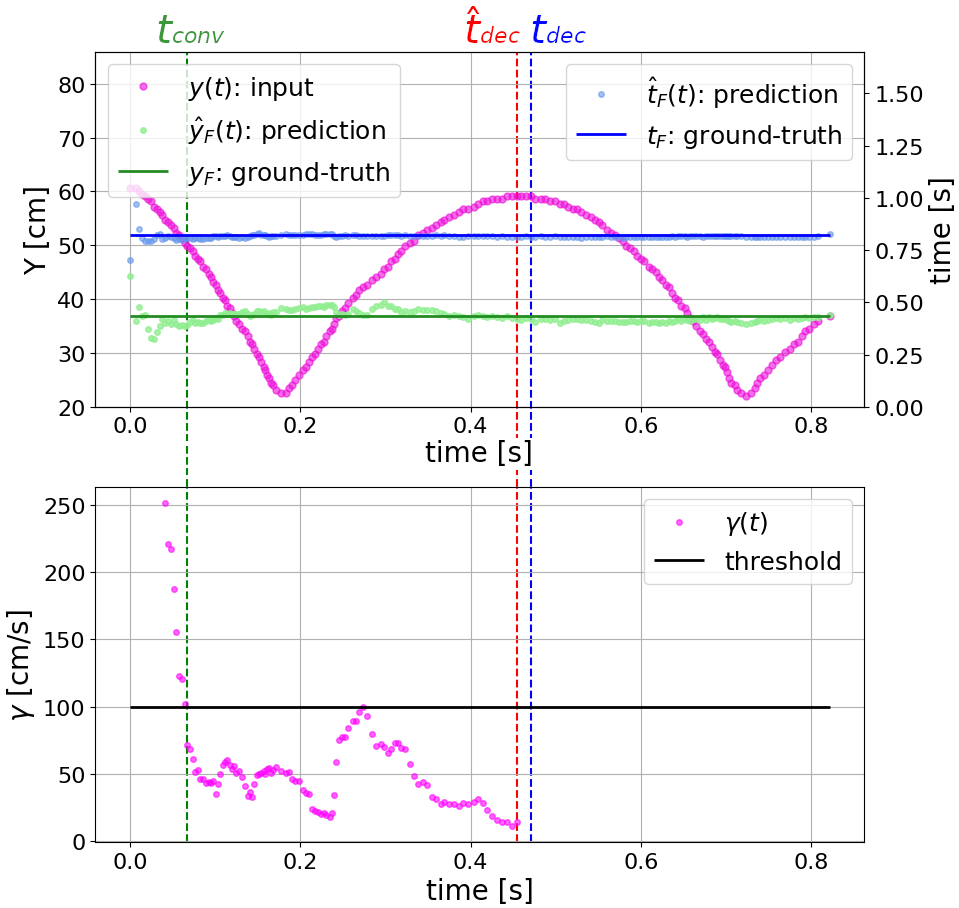}
    \caption{Single trial ground truth and prediction of the final $y$ coordinate of the ball over time: $\gamma$ is the average rate of change of the previous N predictions. The first time $\gamma$ goes under a threshold defines $t_{conv}$, and the last moment the robot can wait to move defines $\hat{t}_{dec}$ (and is determined by the maximum robot's acceleration and distance to the estimated target end point). The robot can act only in the time window between these two instants and if $\gamma$ is below the threshold.}
    \label{fig:params}
\end{figure}

Fig.~\ref{fig:params} shows an example trajectory, the temporal window in which the robot can move (between $t_{conv}$ and $\hat{t}_{dec}$), and $\gamma$.

To fulfill constraints on joint velocities and accelerations of the robot, and to estimate the usefulness of timely and accurate prediction of the end point of the trajectory, the control is implemented in open loop, using only the last position of the target and ignoring subsequent predictions. The robot can then either move as soon as the prediction converges, at $t_{conv}$, to reach the associated $\hat{y}_F$ at $\hat{t}_F$ minimizing the accelerations required; or it moves at the last available moment, $\hat{t}_{dec}$, to possibly gather more information and improve the accuracy of $\hat{y}_F$ and $\hat{t}_F$.

\section{Experiments and Results}
\label{sec:experiments}


To initially compare the event cameras motion-driven sampling with standard frame-like synchronous systems, we train a Stateful LSTM network and carry out the analysis on three types of sampling strategies: event-driven with 2 pixels spatial sampling (\textit{events}), frame-like at 33 Hz (\textit{events33Hz}), and frame-like at 10 Hz (\textit{events10Hz}). 
The term ``\textit{frame-like}'' refers to temporally sub-sampling the full resolution trajectory obtained by the tracker (see Sec.~\ref{subsec:simulated_dataset_generation}), whereby the output of the event-driven tracker is generated at fixed time intervals. This strategy compares the sampling strategies, rather than the full visual pipeline for detection and tracking, that in frame-based cameras might be further affected by motion blur (Fig.~\ref{subfig:setup}).

\subsection{Network training and offline results}
\label{subsec:network_training}
We trained the Stateful LSTM for each sampling strategy with the simulated dataset. The architecture comprises 3 input neurons for the tracker - $(x,y)$ position and the time interval $dt$ between updates - and 2 output neurons that predict the final height $\hat{y}_F$ and time $\hat{t}_F$. Several models were trained for hyperparameters optimization, using the Mean Squared Error (MSE) as loss function. The training was performed using the Adam optimization algorithm with standard parameters, except for the initial learning rate, starting at $\alpha=0.01$, combined with a learning rate decay policy. With these settings, a single hidden layer with 350 neurons was found to be the appropriate network size for all three sampling strategies.

\begin{figure}[!htb]
    \centering
    \begin{subfigure}{.95\linewidth}
        \includegraphics[width=\textwidth]{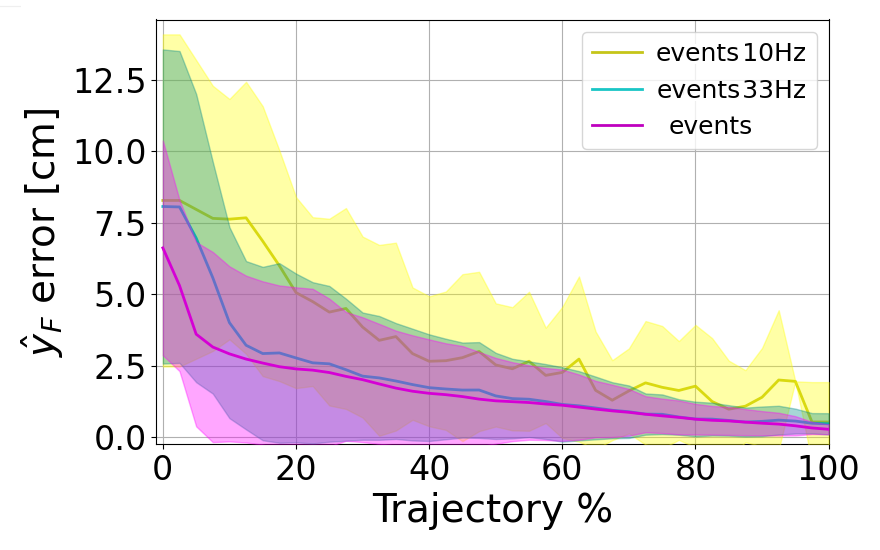}
        \caption{}
        \label{subfig:y_comparison}
    \end{subfigure}
    \begin{subfigure}{.95\linewidth}
        \includegraphics[width=\textwidth]{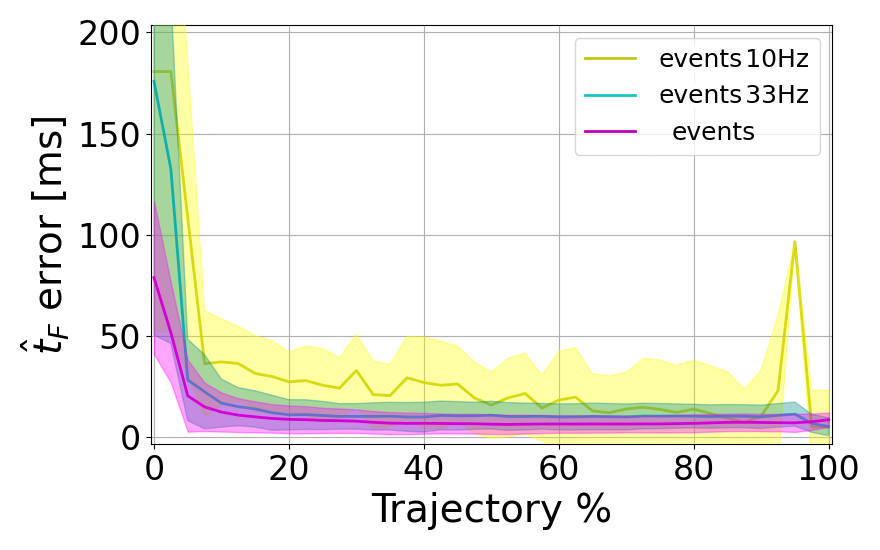}
        \caption{}
        \label{subfig:t_comparison}
    \end{subfigure}
  \caption{Run-time prediction average error (thick line) and standard deviation (band) of the target ending point: (a) Final height $\hat{y}_F$ and (b) final time instant $\hat{t}_F$ prediction error over trajectory execution for the three sampling strategies. Trajectories are normalised (trajectory \%) for comparison as each trajectory has a different number of samples.}
  \label{fig:events_vs_frames}
\end{figure}

Assuming an error threshold of \SI{3.5}{cm} - given by the gripper half-height and ball radius of \SI{2.5}{cm} - the spatial sampling, on average, converged earlier to the correct estimate of the target final position (spatially and temporally) than both temporal sampling methods - as shown in Fig.~\ref{subfig:y_comparison} - increasing the probability of intercepting the ball.

Regarding time prediction, instead, in Fig.~\ref{subfig:t_comparison} the difference is less appreciable since inputting the horizontal spatial coordinate helps estimate the remaining time in any case. Still, a constant offset can be seen between \textit{events} and \textit{events33Hz}.

\begin{figure}[t]
    \centering
    \includegraphics[width=0.45\textwidth]{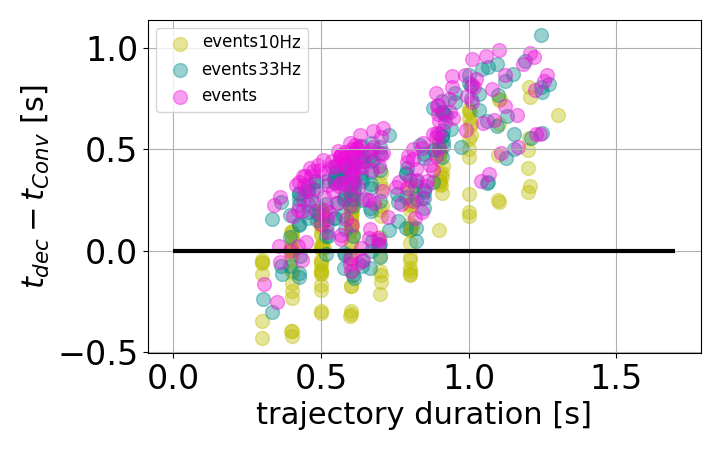}
    \caption{The success of the task is closely linked to the promptness and accuracy of the prediction, through $\hat{t}_{conv}$, but also to the robot velocity, through $\hat{t}_{dec}$. For a trial to be successful, $\hat{t}_{conv}$ must be earlier than $\hat{t}_{dec}$. This condition becomes harder to meet the faster the trajectory and the lower the sampling rate.}
    \label{fig:fixing_robot_speed}
\end{figure}

The faster convergence has benefits in terms of control. Fig.~\ref{fig:fixing_robot_speed} shows how many of the simulated trajectories can be intercepted by moving the Panda robot at $t_{conv}$, given its average velocity, and setting the parameter N of equation \ref{eq:gamma} to 15 for \textit{events}, 3 for \textit{events33Hz}, and 1 for \textit{events10Hz}. The target can be intercepted by the robot only if $t_{conv} < t_{dec}$ and the trajectory mark falls above the black line. The faster the prediction converges, i.e. the smaller $t_{conv}$, the higher the three clouds of samples will be respect to the zero. The same effect can be obtained for increasing $t_{dec}$, using faster robots. 
Samples below the zero black line are trajectories for which the $\hat{y}_F$ prediction did not converge in time to intercept the ball, and their number increases with the speed of the trajectories themselves. They are 10 for the \textit{events} approach, and 19 and 46 for \textit{events33Hz} and \textit{events10Hz}, respectively.

\subsection{Real world experiments}
While the offline prediction accuracy on simulated trajectories establishes a performance baseline, external sources of error like actuation delays or calibration approximations could contribute to the failure of the task. In order to minimise inconsistencies when transferring from the simulated to the real environment, the trained networks have been refined on the real dataset of 310 trajectories. 
We introduce a comparison against a RGB version of our pipeline, with a RealSense D435i camera and a fine-tuned color segmentation algorithm for continuously detecting the ball against the background. We found this option to be the best one among the methods provided by OpenCV, both in terms of accuracy and working principle, compared to our event-driven version. While two different working rates were considered - 30 and 60 Hz - only the latter actually works, due to blur artifacts in 30 Hz images that make the tracking algorithm fail.

\begin{figure}[!t]
    \centering
    \begin{subfigure}{.95\linewidth}
        \includegraphics[width=\textwidth]{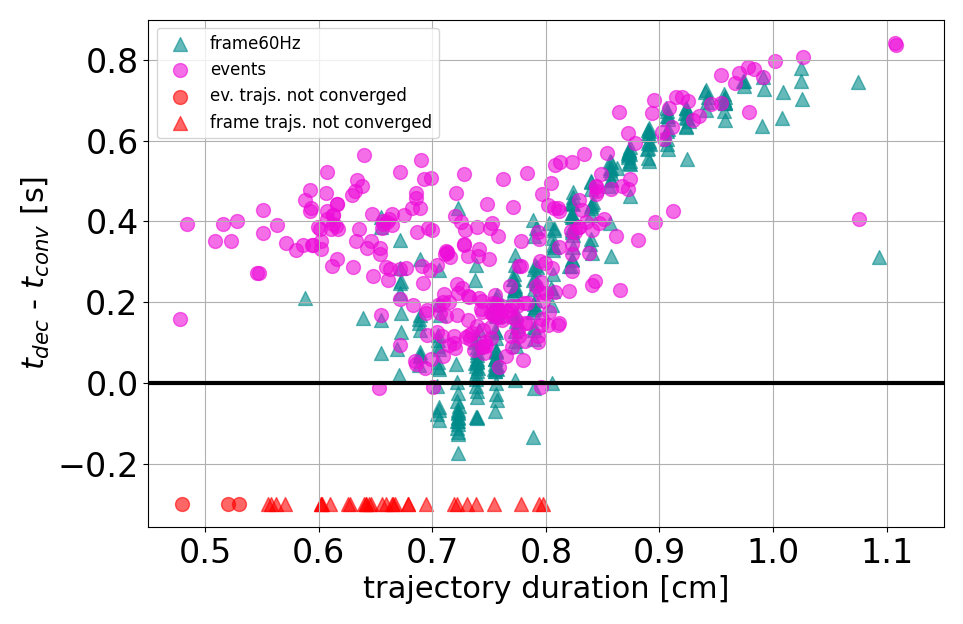}
        \caption{}
        \label{subfig:events_vs_60hz}
    \end{subfigure}
    \begin{subfigure}{.95\linewidth}
        \includegraphics[width=\textwidth]{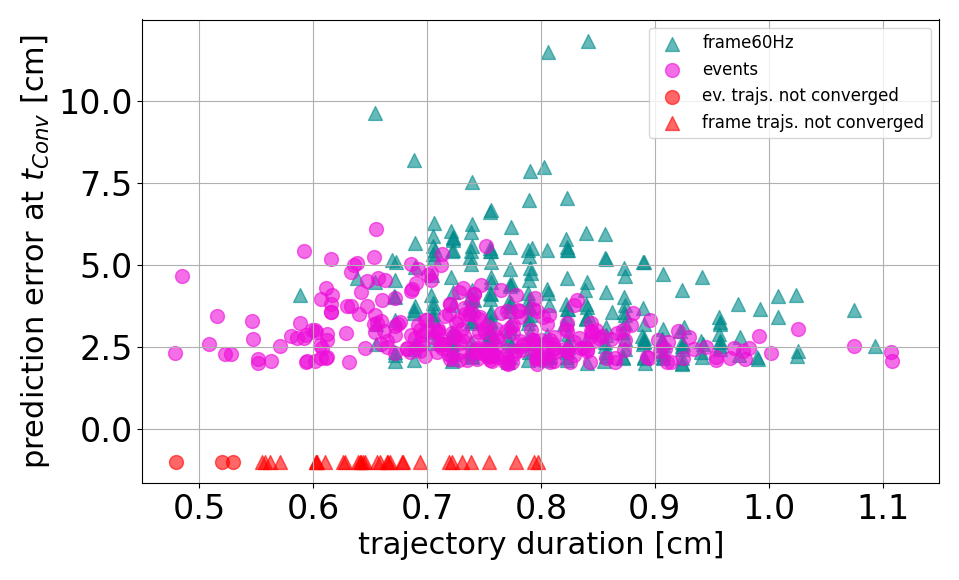}
        \caption{}
        \label{subfig:error_tconv}
    \end{subfigure}
  \caption{(a) Temporal distance between $t_{conv}$ and $t_{dec}$ for the motion-driven sampling strategy of the event camera and the \SI{60}{Hz} time-driven strategy of the RealSense. Values below 0 represent predictions that either converged too late for the robot to move, or did not converge at all (red marks). (b) Prediction error at $t_{conv}$.}
  \label{fig:events_vs_60Hz}
\end{figure}

\begin{figure}[t]
    \centering
    \includegraphics[width=0.45\textwidth]{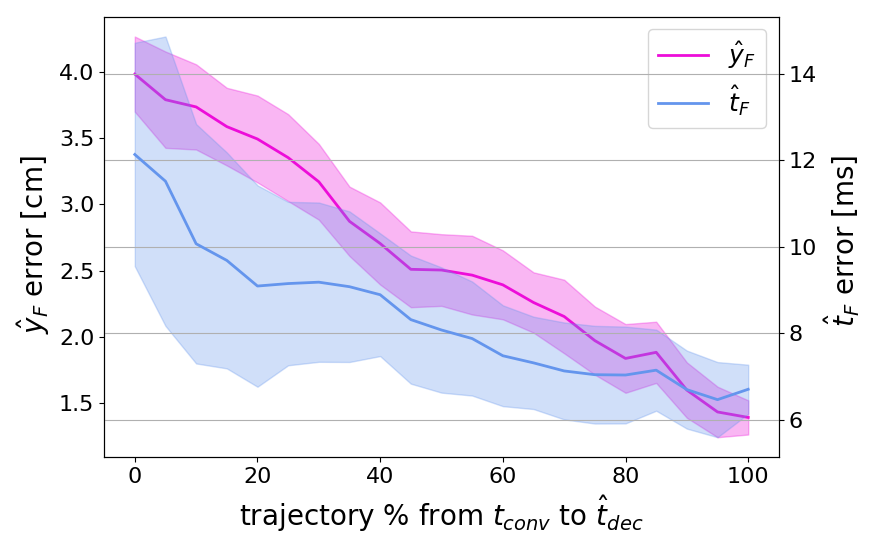}
    \caption{Average error and standard deviation from $t_{conv}$ to $\hat{t}_{dec}$ for the event-driven model. 
    }
    \label{fig:error_after_tconv}
\end{figure}

\begin{figure}[t]
    \centering
    \includegraphics[width=0.45\textwidth]{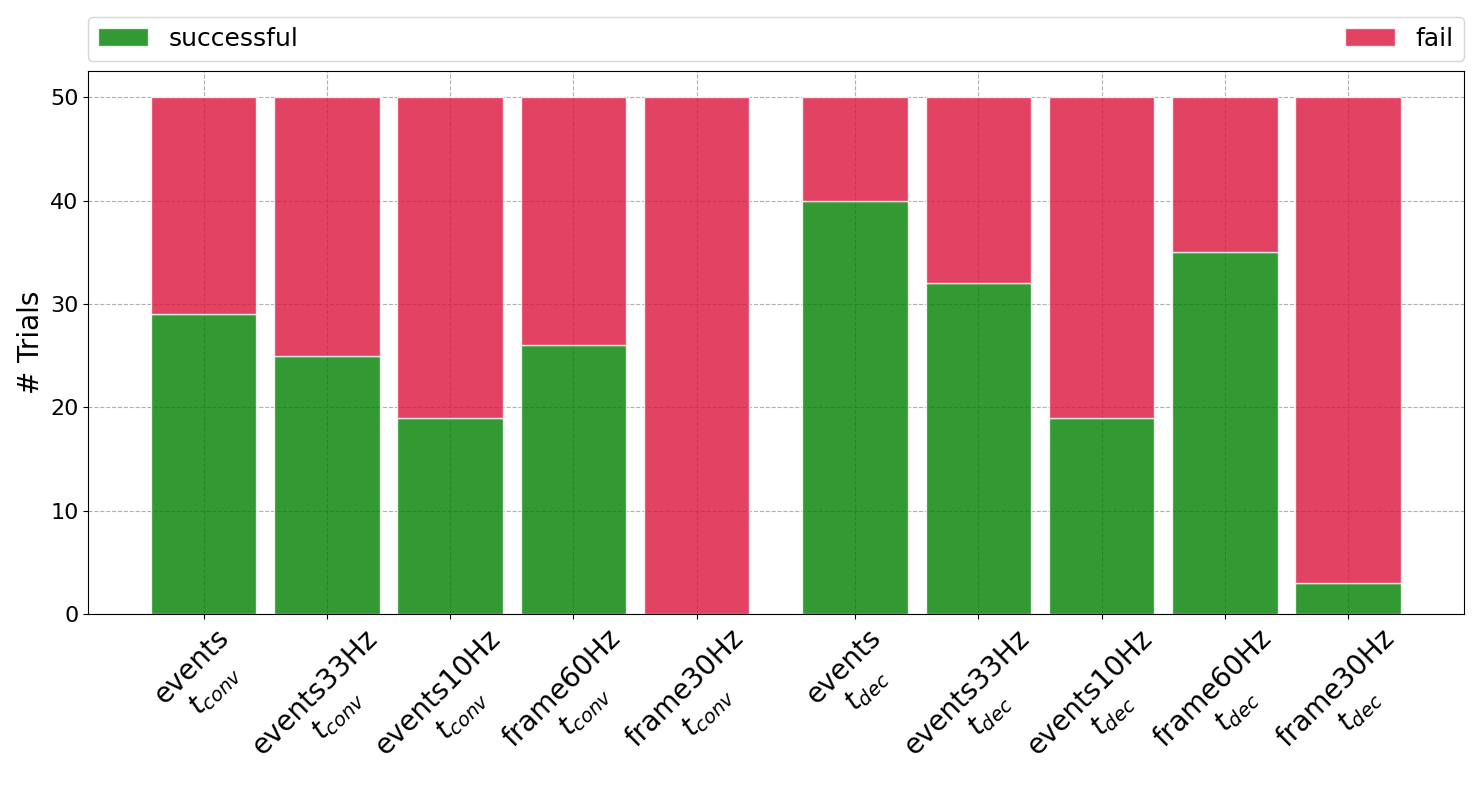}
    \caption{Tracking and prediction performance translates into different numbers of successful trials. Overall, the \textit{events} sampling approach is the most accurate. Moving at $t_{conv}$ leads to a lower accuracy, resulting in a higher number of failures.}
    \label{fig:successful_trials}
\end{figure}

The first analysis compares the motion-driven sampling of event cameras against the highest rate of the RealSense in real applications moving the robot at $t_{conv}$. Fig.~\ref{subfig:events_vs_60hz}, from right to left, shows the difference between $t_{dec}$ and $t_{conv}$.
For trajectories between \SI{0.8}{s} and \SI{1.1}{s} both approaches converge quite in advance, with the gap between $t_{conv}$ and $t_{dec}$ decreasing linearly; for trajectories lasting 0.65-0.8s the event-driven sampling still converges in time (even though with less advance, probably due to the high samples variance in the dataset), while the \SI{60}{Hz} sampling converges too late for 31 samples (samples below the 0) and does not converge at all for 8 of them;
for fast trajectories lasting less than \SI{0.65}{s}, while most of the frame-based cases fail to converge, the event-driven approach keeps converging soon. This is evidence of the fact that, especially for fast trajectories, while a time-driven sampling strategy does not allow to collect enough information for prediction, a motion-driven strategy still gathers enough points to succeed.
The prediction error - shown in Fig.~\ref{subfig:error_tconv} - is consistently lower for the motion-driven approach.
Finally, we run real experiments by throwing the ball at the robot and measuring how many times it was successfully intercepted, for the five different sampling strategies and when the movement of the robot is triggered at the two possible times of action: $t_{conv}$ or $\hat{t}_{dec}$. A total of 500 trials were run, 50 for for each sampling strategy and time of action.
In the $t_{conv}$ case, the \textit{events} model, given the high temporal resolution, captures most of the trajectory and shows the highest number of hit rates, though \textit{events33Hz} and \textit{frames60Hz} are comparable.
Moving the robot at $\hat{t}_{dec}$, instead, pushes the robot actuation to its limits, but also increases the success rate of the three strategies, thanks to the more information accumulated, as shown in Fig.~\ref{fig:error_after_tconv}.
For the \textit{events10Hz} case, instead, this is not true, probably since, while the neural network predicts correctly, the too-low sampling rate does not allow to determine the exact height at which the ball goes out of the field of view. In other words, if the last visible sample happens \SI{90}{ms} before the trajectory ends, the prediction will converge to this value, which does not represent the real ending point. Same holds, in a more limited way, for the \textit{events33Hz} strategy.
Regarding the \textit{frame33Hz} case, finally, we reported the data only for completeness. Due to image blur caused by the speed of the ball, the tracker was not able to segment the ball, leading to complete failure in the $t_{conv}$ case and to very few successes in the $\hat{t}_{dec}$ case, when the trajectory was slow enough to be followed. Results are summarized in Fig.~\ref{fig:successful_trials}.


\section{Conclusions}
\label{sec:conclusions}


We investigated the advantages of using an event camera for predicting the end-point of a target's trajectory to be intercepted with a robotic arm. We did not use a specific model of the task, to have a more general approach applicable to other use cases. We adopted a fully data-driven Stateful LSTM network, exploiting the possibility to send one sample at a time, without restrictions on the number of input points. The state of the network is updated with each observation without the need for buffering and defining a priori the length of redundant input. To bootstrap the learning process, we resorted to a dataset of simulated trajectories, generated in Unreal Engine, and refined the models on real data.
Differently from~\cite{Wang2022}, our system is purely event-driven, avoiding any fixed-rate time-driven representation like BEHI, and does not need a dedicated GPU for processing, but smoothly runs on an Intel i7-9750H CPU.

Comparing the event approach with a 2 pixels spatial sampling against two fixed-rate time-driven samplings of \SI{33}{Hz} and \SI{10}{Hz} shows the benefit of the motion-driven principle in faster convergence to a lower error. This translates into a higher rate of successful ball interception compared to the two frame-like sampling strategies. 
Sampling the trajectory at too low rates, moreover, can result in missing relevant information about the end-point, as it happened for the \SI{10}{Hz} case. 
On the other hand, from the experiments at \SI{60}{Hz} with a RealSense camera, it appears that similar results to event cameras can be achieved with a higher enough frame rate. Nevertheless, such a system would still require the continuous acquisition of the whole image, specific illumination conditions related to the higher shutter speed of the camera required to decrease the effect of motion blur, and, overall, more computational power. On the other hand, the 2 pixels spatial sampling can be increased to have a lower rate, if necessary, but still preserving the motion-driven principle. 
Overall, we believe the event-driven approach could be beneficial for applications characterized by short objects' in-flight times and fast robotic avoidance maneuvers of obstacles with trajectories that are not easily parameterizable.

In our study case, moving the robot as soon as the prediction converges still brings with it an error affecting the success of the task, while waiting for the last moment requires faster movements, but leads to higher precision and success rate. To improve on this aspect, depth could be included to expand the operative space, and a third close-loop control strategy could be implemented, starting to move the robot as soon as the prediction converges but updating the desired final position of the movement with the refinement of the prediction obtained with additional points along the trajectory. 
An interesting direction is the introduction of physics into the network to further push convergence speed and accuracy, at the cost of tailoring to the task at hand. Similarly to~\cite{Asenov2020}, this might be achieved by introducing an additional training loss term to encode physical constraints like gravity and elastic impact equations.







\bibliographystyle{IEEEtran}
\bibliography{bibliography.bib}

\end{document}